\documentclass{article}

\usepackage[dblblindworkshop,final]{neurips_2026}

\workshoptitle{Generative and Experimental Perspectives for Biomolecular
Design}

\usepackage[utf8]{inputenc}
\usepackage[T1]{fontenc}
\usepackage{amsmath,amssymb,amsthm}
\usepackage{booktabs}
\usepackage{graphicx}
\usepackage{float}
\usepackage{url}
\usepackage[hidelinks]{hyperref}
\usepackage{microtype}

\newcommand{\AgoHi}{170,000}
\newcommand{\AgoLo}{15,000}

\newcommand{\AlphaThreshLo}{0.26}
\newcommand{\ApxB}{100}
\newcommand{\ApxCases}{54}
\newcommand{\ApxCostElems}{1.25}
\newcommand{\ApxCostSecond}{1.63}
\newcommand{\ApxDefectMed}{\ensuremath{1.7\times 10^{-5}}}
\newcommand{\ApxDefectWorst}{\ensuremath{2.0\times 10^{-3}}}
\newcommand{\ApxDevFam}{8}
\newcommand{\ApxElems}{4,819}
\newcommand{\ApxFDAbove}{13}
\newcommand{\ApxFDCoord}{17}
\newcommand{\ApxFDStrat}{0.00144}
\newcommand{\ApxFDWorst}{0.0433}

\newcommand{\ApxFullMs}{7.3}
\newcommand{\ApxGrad}{0.0047}
\newcommand{\ApxGuideOverlap}{0}
\newcommand{\ApxIters}{200}
\newcommand{\ApxLinGrad}{\ensuremath{1.8\times 10^{4}}}
\newcommand{\ApxMethod}{bins}
\newcommand{\ApxNRef}{4,919}
\newcommand{\ApxR}{100}
\newcommand{\ApxReps}{50}
\newcommand{\ApxResid}{\ensuremath{2.7\times 10^{-16}}}
\newcommand{\ApxRhos}{6}
\newcommand{\ApxSeedOverlap}{0}
\newcommand{\ApxSelMs}{15.3}
\newcommand{\ApxTestCon}{9}
\newcommand{\ApxTestFam}{7}
\newcommand{\ApxTruncGrad}{1.00}

\newcommand{\BSmallestOne}{100}

\newcommand{\BestInv}{0.00280}
\newcommand{\BestInvCI}{[0.00080,\, 0.00470]}
\newcommand{\BestMonotone}{not monotone}
\newcommand{\BestNInv}{17}
\newcommand{\BestNTested}{24}

\newcommand{\BudgetAtAgoLo}{0.29}

\newcommand{\BudgetWorst}{1,491}
\newcommand{\CanEightMerCI}{[-0.01960,\, -0.01420]}
\newcommand{\CanEightMerMed}{-0.01720}
\newcommand{\CanEightMerN}{2,357}
\newcommand{\CanInv}{0.00305}
\newcommand{\CanInvCI}{[0.00115,\, 0.00478]}
\newcommand{\CanMonotone}{not monotone}
\newcommand{\CanNInv}{16}
\newcommand{\CanNTested}{24}
\newcommand{\CanSevenAOneCI}{[-0.01190,\, -0.00915]}
\newcommand{\CanSevenAOneMed}{-0.01050}
\newcommand{\CanSevenAOneN}{4,860}
\newcommand{\CanSevenMEightCI}{[-0.00860,\, -0.00640]}
\newcommand{\CanSevenMEightMed}{-0.00745}
\newcommand{\CanSevenMEightN}{9,955}
\newcommand{\CanSixMerCI}{[-0.00410,\, -0.00240]}
\newcommand{\CanSixMerMed}{-0.00315}
\newcommand{\CanSixMerN}{17,504}

\newcommand{\ClassSane}{5}
\newcommand{\ClassSlope}{1.42}

\newcommand{\ClassSlopeCI}{[0.96,\, 2.52]}

\newcommand{\ClusterBoot}{2000}

\newcommand{\CtxAgree}{3}
\newcommand{\CtxAttempted}{7}
\newcommand{\CtxDeliveryMax}{1.89}
\newcommand{\CtxDeliveryMin}{0.72}
\newcommand{\CtxExcl}{2}

\newcommand{\CtxRatio}{0.44}
\newcommand{\CtxRatioCI}{[0.32,\, 0.61]}

\newcommand{\DoseKMaxAbsSpearman}{0.079}
\newcommand{\DoseLevels}{1, 10, 25}

\newcommand{\DoseTwoName}{HK2-4031}

\newcommand{\EFourAbundPct}{81.9}
\newcommand{\EFourAff}{91.1}

\newcommand{\EFourN}{952}

\newcommand{\EFourRShallow}{25}

\newcommand{\GapClusterCI}{[-0.0337,\, -0.0051]}

\newcommand{\GapNet}{0.0026}
\newcommand{\GapNetOverSE}{0.22}

\newcommand{\GapNullMean}{-0.0242}
\newcommand{\GapRho}{0.100}
\newcommand{\GencodeRelease}{50}
\newcommand{\GradErr}{\ensuremath{2.3\times 10^{-9}}}
\newcommand{\GradWorst}{\ensuremath{1.6\times 10^{-4}}}
\newcommand{\GradWorstNorm}{\ensuremath{2.3\times 10^{-7}}}

\newcommand{\LocoHi}{1.77}
\newcommand{\LocoLo}{0.63}

\newcommand{\MaxConstructsPerGene}{23}

\newcommand{\MaxNetGapWhereMet}{\ensuremath{4.0\times 10^{-4}}}
\newcommand{\MaxNetOverSE}{0.27}
\newcommand{\NArraysHela}{67}
\newcommand{\NBelowOne}{3}
\newcommand{\NBinSettings}{20}
\newcommand{\NBirmArrays}{24}

\newcommand{\NBirmGuides}{12}
\newcommand{\NCanonUTR}{19,786}
\newcommand{\NConstructs}{25}
\newcommand{\NConstructsScored}{24}

\newcommand{\NCritMet}{3}

\newcommand{\NCtxArrays}{56}

\newcommand{\NDoseArrays}{54}

\newcommand{\NDoseFitted}{5}

\newcommand{\NDoseThree}{4}

\newcommand{\NGapCIExclZero}{5}
\newcommand{\NGeneRepeat}{8,068}
\newcommand{\NGradCoord}{35}

\newcommand{\NMSlopeExclOne}{4}
\newcommand{\NMSlopeTested}{5}

\newcommand{\NPrefFree}{5}

\newcommand{\NSeedVariants}{19}
\newcommand{\NThermoNeg}{3}
\newcommand{\NmRNA}{80,000}

\newcommand{\NullCentredEq}{-0.0540}
\newcommand{\NullCentredInd}{-0.0567}
\newcommand{\NullConstructs}{24}
\newcommand{\NullNPairs}{34,676}
\newcommand{\NullNRho}{6}

\newcommand{\NullPerm}{1000}
\newcommand{\NullZHi}{-10.5}
\newcommand{\NullZLo}{-11.5}
\newcommand{\PwDecades}{9}
\newcommand{\PwPredicted}{-2}
\newcommand{\PwSlope}{-2.00}
\newcommand{\PwSlopeAll}{-0.99}
\newcommand{\PwSlopeAllSE}{0.00}
\newcommand{\PwSlopeMax}{-1.00}
\newcommand{\PwSlopeMaxSE}{0.001}
\newcommand{\PwSlopeSE}{0.002}
\newcommand{\QuantileCheaperClause}{never the cheaper rule}
\newcommand{\RedistDraws}{1200}

\newcommand{\RedistViolA}{0}
\newcommand{\RedistViolB}{0}
\newcommand{\RegDiff}{0}

\newcommand{\RhoDataDecades}{4.8}

\newcommand{\RhoDecades}{4.1}

\newcommand{\RhoHi}{2.1}
\newcommand{\RhoLo}{\ensuremath{1.9\times 10^{-4}}}

\newcommand{\SimKErr}{3.2}
\newcommand{\SimNoise}{0.05, 0.15, 0.3}
\newcommand{\SimPoolErr}{4.1}
\newcommand{\SimSlopeErrClass}{0.0056}
\newcommand{\SimSlopeErrEq}{0.0055}
\newcommand{\SimSlopeErrInd}{0.019}
\newcommand{\SimTrueEqSlope}{2.423}

\newcommand{\SpearBest}{-0.0746}

\newcommand{\SpearBestPairCI}{[-0.0854,\, -0.0643]}
\newcommand{\SpearBestRho}{0.100}
\newcommand{\SpearGap}{-0.0216}
\newcommand{\SpearInd}{-0.0530}

\newcommand{\TauContour}{0.044}

\newcommand{\ThermoSlope}{1.02}
\newcommand{\ThermoSlopeCI}{[-0.15,\, 1.81]}
\newcommand{\WithinDiff}{\ensuremath{1.7\times 10^{-5}}}

\newtheorem{proposition}{Proposition}

\title{One pool, many targets: a conservation layer and what
archival data can identify}

\author{%
  Shadi Khodagholi \\
  Department of Industrial Engineering \\
  University of Central Florida \\
  \texttt{zahra.khodagholi@ucf.edu}
  \And
  Niloofar Yousefi \\
  Department of Industrial Engineering \\
  University of Central Florida \\
  \texttt{niloofar.yousefi@ucf.edu}
}

\begin{document}
\maketitle

\begin{abstract}
Pairwise guide--transcript scores do not enforce conservation of a finite
guide-loaded RISC pool when they are interpreted independently as occupancies.
We formulate a differentiable scalar equilibrium layer: one conservation
equation with a unique positive root and exact implicit gradients. It yields
a redistribution theorem, a qualified high-resource limit, an analysis of the
retrieval approximation, and a conditional rank-invariance result: within one
construct at one dose, rankings by fractional occupancy cannot distinguish
equilibrium from independent scoring. We therefore audit the two experiments
that proposition leaves open, dose and cross-context, on archival off-target
data. Corrected thermodynamic affinities associate weakly with measured
repression in the direction a working predictor requires, but a paired
permutation test and a construct-cluster bootstrap do not establish added
predictive value from the coupling: what survives their differing
permutation-null baselines is \GapNet{}, a descriptive \GapNetOverSE{} of the
equilibrium association's cluster standard error.
The dose fits are heterogeneous and frequently violate the model-implied
exponent constraint, which is superlinear rather than sublinear, so these data
do not identify the competition parameter. A saturable compression of the
competitor set holds both accuracy targets on held-out guide families but is
not faster at the size measured. The contribution is a reusable conservation
operator and the experimental information needed to test it. The code for this study is available at
\url{https://github.com/shadi97kh/One-Pool-Many-Targets}.
\end{abstract}

\section{Introduction}

A transfected siRNA silences its target and, through its seed, tens to
hundreds of unintended transcripts \citep{jackson2003, birmingham2006,
jackson2006seed}. Design tools score candidates with sequence models
\citep{huesken2005, bai2024} that evaluate each (guide, transcript) pair
independently, which leaves the shared loaded-RISC budget unenforced. Cells
supply that constraint: Argonaute has limited copy number \citep{wang2012, janas2012},
transfected small RNAs compete with endogenous microRNAs for it
\citep{khan2009}, and target abundance dilutes activity \citep{arvey2010,
bosson2014, denzler2014, denzler2016}. We model one part of this: competition
among target sites for a pool \emph{already} loaded with the candidate guide.
Loading competition, endogenous microRNA occupancy, strand selection,
recycling and cleavage enter only through $M$.

Quantitative models of small-RNA competition already exist
\citep{loinger2012}, as do explicit on- and off-target repression models for
screening data \citep{riba2017}. Ours is narrower and mostly structural: the
operator, its comparative statics, and an audit of what archival data can
identify.

\section{A shared-resource equilibrium layer}
\begin{figure}[t]
\centering
\includegraphics[width=\textwidth]{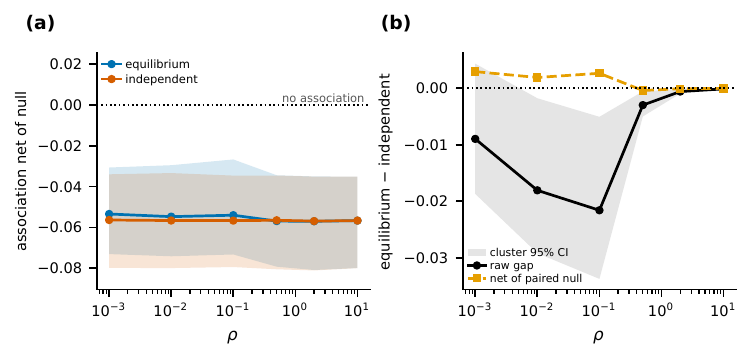}
\caption{Does the shared-pool coupling add anything over independent
scoring? GSE5814, \NullConstructs{} constructs, \NullNPairs{} pooled
construct--transcript pairs. (a) Pooled Spearman against measured log ratio,
each scoring net of \emph{its own} within-construct permutation null
(\NullPerm{} draws); bands are \ClusterBoot{} construct-cluster 95\%
intervals. Repression is negative, so below zero is the working direction.
(b) Their difference. The raw gap is reliably nonzero, its cluster interval
excluding zero at \NGapCIExclZero{} of \NullNRho{} values of $\rho$, yet under
a paired null (one shuffle applied to both models) what survives is at most
\MaxNetOverSE{} of one cluster standard error (\S\ref{sec:assoc}).}
\label{fig:assoc}
\end{figure}

Let $M$ be the pool of RISC loaded with the candidate guide, $x_j$ the
abundance of transcript $j$, and $K_j>0$ an effective interaction parameter.
Writing $f$ for the free loaded pool, mass conservation over the retrieved
set $\mathcal{R}$ gives
\begin{equation}
F(f) \;=\; (1+\beta)\,f \;+\; \sum_{j\in\mathcal{R}} \frac{x_j f}{K_j+f}
\;-\; M \;=\; 0,
\qquad
o_j \;=\; \frac{x_j f}{K_j + f},
\label{eq:cons}
\end{equation}
where a constant $\beta$ stands in for the omitted set as a linear,
unsaturated reservoir, which cannot represent saturable omitted targets
exactly (\S\ref{sec:binned}). Combining a transcript's sites as
$1/K_j = \sum_s 1/K_{js}$ and capping its load at $x_j$ treats them as
\emph{mutually exclusive}: at most one complex per transcript
(\S\ref{sec:proofs}). Since $F(0)=-M<0$,
$F'>0$ and $F(M)\ge 0$, the root is unique and bracketed by $[0,M]$, so
bisection is globally convergent and batches over candidates; gradients come
from the implicit function theorem and match central differences to a median
relative error of \GradErr{} (coordinate-level checks, \S\ref{sec:proofs}).

The independent counterfactual is the same expression with $f$ pinned to $M$,
$o_j^{\text{ind}} = x_j M/(K_j+M)$, isolating the coupling as the only
difference. The competition parameter is
$\rho = M/N_{\mathrm{mRNA}}$, where the denominator is the \emph{whole-cell}
mRNA count, \NmRNA{} molecules per cell, and not the retrieved set
$\mathcal{R}$ of \eqref{eq:cons}. That is what makes $\rho$ the same
quantity as $\alpha\cdot(\text{Argonaute})/(\text{mRNA})$ in
\S\ref{sec:regime}.
The pipeline is short: control-channel abundances and GENCODE 3$'$UTRs give
$x_j$, a ViennaRNA construction anchored on a measured $K_d$ gives $K_j$, and
\eqref{eq:cons} turns them into occupancies whose gradients flow back to
$K$ (\S\ref{sec:workflow}, Figure~\ref{fig:workflow}).
Whether independent scoring over-allocates is conditional on the loaded
fraction $\alpha$: at \AgoLo{} total Ago1--4 in HeLa with $\alpha=1$ it uses
\BudgetAtAgoLo{} of the budget, crossing only below
$\alpha=\AlphaThreshLo{}$ (Figure~\ref{fig:mech}).

\subsection{What the layer implies}

\paragraph{Redistribution, the high-resource limit, retrieval depth.}
Weakening one transcript's affinity frees pool for every other,
$\partial o_i/\partial K_t > 0$ for $i \neq t$ and $\partial o_t/\partial
K_t<0$, with \RedistViolA{} and \RedistViolB{} violations over \RedistDraws{}
draws (\S\ref{sec:proofs}). With $\beta=0$ the fractional-occupancy gap is
$O(M^{-2})$ pointwise as $M\to\infty$, so equilibrium contains independent
scoring as its high-resource limit under conditions in \S\ref{sec:limit}.
Shallow truncation is unusable: at $R=\EFourRShallow{}$ affinity errs by
\EFourAff\% and abundance mass by \EFourAbundPct\% (\S\ref{sec:binned}).

\subsection{Compressing the competitor set}
\label{sec:approx}

\begin{figure}[t]
\centering
\includegraphics[width=\textwidth]{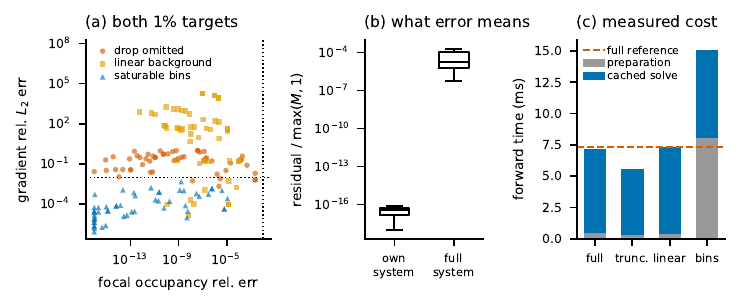}
\caption{Compressing the competitor set: \ApxCases{} held-out cases,
\ApxTestFam{} guide families disjoint from the \ApxDevFam{} used to fix the
configuration. (a) Relative error against the full retrieved reference at
$R=\ApxR{}$, one point per case, three treatments of the omitted transcripts;
dotted lines are the 1\% targets, so only the lower-left quadrant passes both.
(b) Two error measures for the same solutions, each divided by $\max(M,1)$;
median and quartiles. (c) Median forward time (ms) over \ApxNRef{}
competitors, single-threaded float64; bin construction is part of
preparation.}
\label{fig:approx}
\end{figure}

Truncation is usable only if the omitted mass is represented well enough to
differentiate through. At depth $R=\ApxR{}$ we compare three treatments of the
remainder: discard it, fold it into one linear reservoir, or bin it into
\ApxB{} saturable groups, $X_b=\sum x_j$ and $K_b = X_b/\sum(x_j/K_j)$. The
configuration was fixed on \ApxDevFam{} guide families from D1 and D2, with
1\% relative error in the focal occupancy \emph{and} the gradient as the prior
criterion,
then applied unchanged to \ApxTestFam{} disjoint families, \ApxCases{} cases.

Only the saturable bins pass both (Figure~\ref{fig:approx}a): worst gradient
error \ApxGrad{} against \ApxTruncGrad{} for discarding and \ApxLinGrad{}
for the linear reservoir, though all three are inside 1\% in the forward
solve. A linear background has constant slope in $f$, so once the omitted mass
saturates it reports the wrong $\partial F/\partial f$, which implicit
gradients divide by. Accuracy is the defect against the \emph{original}
conservation law, median \ApxDefectMed{} and worst \ApxDefectWorst{}; the
compressed system's own residual, \ApxResid{}, says only that the bisection
converged (Figure~\ref{fig:approx}b). It is not faster: \ApxSelMs\,ms
forward against \ApxFullMs\,ms for the full reference at \ApxNRef{}
competitors, because the solver runs a fixed iteration count and a second
reduction costs more than a longer one at this size
(Figure~\ref{fig:approx}c, \S\ref{sec:apxdetail}). Accurate, with no measured
speed-up.

\paragraph{A negative result that constrains within-construct fixed-dose
rank tests based on fractional occupancy in the one-effective-$K$
formulation.}

\begin{proposition}[Within-construct fractional-occupancy rank invariance]
\label{prop:rank}
Fix a construct and a dose, and represent each transcript by one effective
binding state, so its sites are mutually exclusive and it carries at most one
complex. Equilibrium fractional occupancy is $q^{\mathrm{eq}}_j=f/(K_j+f)$ and
independent fractional occupancy is $q^{\mathrm{ind}}_j=M/(K_j+M)$. Both $f$
and $M$ are scalars shared by every transcript, and both expressions are
strictly decreasing in $K_j$. Hence one is a strictly monotone transform of
the other, and every rank statistic over transcripts computed on fractional
occupancy is identical, at every $\rho$.
\end{proposition}

The scope is exactly that: not absolute occupancy $x_j q_j$, which reorders
whenever abundance differs, not simultaneously occupiable sites. Within it the
result is a property of the model, not of our data, and it disqualifies the
experiment we first set out to run; the largest within-construct Spearman
difference over all $\rho$ is \WithinDiff{}, at the solver's bisection
tolerance. The layer can reorder only across constructs, doses or
transcriptomes.

\section{Data}
\label{sec:data}

Archival off-target microarrays only, numbered in order of appearance
(\S\ref{sec:recover} maps these onto the repository's D1--D9). \textbf{D1},
GSE5814 \citep{jackson2006seed}: \NArraysHela{} HeLa arrays, \NConstructs{}
constructs including \NSeedVariants{} seed variants of one MAPK14 siRNA. One is the non-targeting luciferase
control, which has no human guide, so \NConstructsScored{} are scored.
\textbf{D2}, E-MEXP-668 \citep{birmingham2006}: \NBirmArrays{} HeLa arrays,
\NBirmGuides{} guides, recovered from ArrayExpress
(\S\ref{sec:recover}); its seeds supply most of both sides of the
\S\ref{sec:approx} split and share \ApxGuideOverlap{} exact guides and
\ApxSeedOverlap{} exact seeds (2--8) with D1. \textbf{D3}, dose series GSE28786 \citep{caffrey2011}:
\NDoseArrays{} arrays. \NDoseThree{} constructs were measured at
\DoseLevels\,nM and \DoseTwoName{} only at 1 and 10\,nM; matched 0\,nM
controls were also deposited.
\textbf{D4}, cross-context GSE14073 \citep{burchard2009}: \NCtxArrays{}
arrays in two liver lines. Recovery and preprocessing: \S\ref{sec:recover}.

Transcript structure is GENCODE v\GencodeRelease{} \citep{frankish2021},
\NCanonUTR{} canonical 3$'$UTRs. Abundance comes from the untreated, mock or
non-targeting control measurements of the same experiment: the reference
channel for the two-colour GSE5814 and E-MEXP-668 arrays, and the
control-condition arrays for the single-channel GSE28786 and GSE14073. Affinity is thermodynamic,
$\Delta G_{\mathrm{eff}} = \Delta G_{\text{duplex}} - RT\ln
P_{\text{unpaired}}$ from ViennaRNA \citep{lorenz2011, bernhart2006}, the
duplex energy \emph{plus} the cost of opening the site, anchored on the
measured seed-match $K_d$ \citep{wee2012} to give molecules per cell
(\S\ref{sec:scale}); $K$ is never fitted to expression. Argonaute copy number
\citep{wang2012, janas2012} is carried as a range; the mRNA total
\citep{marinov2014} is one value inside the range that measurement reports
(\S\ref{sec:limits}).

\section{Where a real transfected cell sits}
\label{sec:regime}

With abundance normalised to the assumed total, $\rho_{\mathrm{cell}}$
reduces to $\alpha \cdot (\text{Argonaute copies})/(\text{mRNA copies})$,
$\alpha$ being the unmeasured loaded fraction. The two Argonaute figures are
not two measurements of one quantity: \AgoLo{} is total Ago1--4 in HeLa and
\AgoHi{} total Argonaute from other settings, neither AGO2-specific, and the
mRNA count was not measured on these arrays, so what follows is a
literature-based \emph{scenario range}, not an estimated band. Sweeping
$\alpha$ over three decades, with the Argonaute figures spanning a further
decade, gives $\rho_{\mathrm{cell}} \in [\RhoLo, \RhoHi]$, and the contour
below which $\tau<0.9$ sits at $\rho \approx \TauContour{}$, \emph{inside} it
(Figure~\ref{fig:phase}). Whether the layer matters therefore turns on a
quantity these datasets do not measure; whether a dose series can supply it is
\S\ref{sec:dose}.

\section{What the affinity scores predict}
\label{sec:assoc}

Over \NullNPairs{} pooled construct-transcript pairs in D1 the equilibrium
score reaches \SpearBest{} at $\rho=\SpearBestRho{}$ against \SpearInd{} for
independent scoring, each \NullZLo{} to \NullZHi{} standard deviations from
its own permutation null (Figure~\ref{fig:assoc}a): weak, and in the right
direction.

The raw difference between them is not evidence that coupling helps: the two
have different null baselines, because pooling constructs leaves a
between-construct correlation a within-construct shuffle cannot remove. At
$\rho=\GapRho{}$, where they differ most, the raw gap is \SpearGap{} but the
paired null, one shuffle applied to both models, already sits at
\GapNullMean{} and so exceeds it. What survives is \GapNet{}, or
\GapNetOverSE{} of the equilibrium association's cluster standard error: a
descriptive scale, not a paired test. The prespecified three-part screen of
\S\ref{sec:paired} is met at \NCritMet{} of \NullNRho{} evaluated $\rho$
values, all at the largest $\rho$, where the scorings converge by construction
and the absolute difference is at most \MaxNetGapWhereMet{}; the separate
practical-significance requirement is met at no $\rho$. This analysis does not
establish that coupling improves prediction.

Per dose the correlation is at most \DoseKMaxAbsSpearman{} in absolute value,
so this $K$ orders transcripts only coarsely: 7mer-A1 sites are repressed more
than 7mer-m8, a $t1$-pocket contribution \texttt{duplexfold} does not
represent (\S\ref{sec:hier}).

\begin{figure}[t]
\centering
\includegraphics[width=0.78\textwidth]{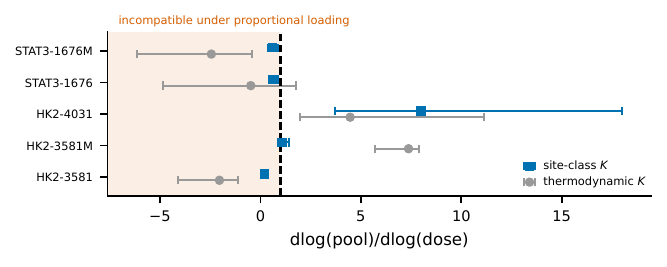}
\caption{Fitted dose exponent
$\mathrm{d}\log(\text{pool})/\mathrm{d}\log(\text{dose})$ per construct on
GSE28786, both estimators, 95\% bootstrap intervals, \NDoseFitted{}
constructs. Conservation makes the pool convex in the loaded amount, so under
proportional loading the exponent cannot fall below $1$: the shaded region
indicts that combined specification rather than equilibrium alone. The
constructs are variants of one another, so the audit cannot separate mechanism
from assumption (\S\ref{sec:convex}).}
\label{fig:dose}
\end{figure}

\section{A dose-series identifiability audit}
\label{sec:dose}

Proposition~\ref{prop:rank} fixes the dose; it says nothing about how one
construct behaves across doses, because dose changes $M$ and therefore $f$.
\citet{caffrey2011} profile \NDoseThree{} of \NDoseFitted{} constructs at
\DoseLevels\,nM and \DoseTwoName{} at 1 and 10\,nM only, against matched 0\,nM
arrays, holding \emph{total} transfected duplex constant. We \emph{assume} the
guide-specific loaded pool $M$ is proportional to administered dose: the
constant-duplex design makes that more plausible but does not establish
proportional uptake, loading or recycling. We fit one scalar pool per dose by
maximum likelihood with $K_j$ fixed, regress $\log(\text{pool})$ on
$\log(\text{dose})$, then check the assumption.

\paragraph{The direction of the test is the opposite of what we expected.}
One expects added complex to be absorbed, so an exponent below $1$ would
signal competition. Conservation says otherwise: $M(f)$ is concave in $f$, so
$f$ is \emph{convex} in $M$ and $\mathrm{d}\log f/\mathrm{d}\log M \ge 1$.
The measured exponent is against \emph{dose}, and inherits that bound only
under proportional loading, so an exponent below $1$ indicts the combined
specification rather than equilibrium (\S\ref{sec:convex}).

We report two estimators. The thermodynamic $K$ gives a pooled exponent of
\ThermoSlope{} (95\% CI \ThermoSlopeCI) with \NThermoNeg{} of \NDoseFitted{}
per-construct exponents \emph{negative}, which conflict with the fixed-input
model whenever the loaded pool is nondecreasing in dose, proportional loading
included (\S\ref{sec:convex}), so it diagnoses its own failure. The second carries
affinity by \emph{site class}, one parameter per class shared across doses
(\S\ref{sec:sim}).

\paragraph{The audit fails.} The site-class pooled exponent is \ClassSlope{}
with bootstrap interval \ClassSlopeCI{} containing $1$, but it is not robust:
leaving each construct out moves it over \LocoLo--\LocoHi{}, and
\NBelowOne{} of \NDoseFitted{} construct-level intervals lie entirely below
$1$, incompatible with the combined model under proportional loading
(Table~\ref{tab:e9}, Figure~\ref{fig:dose}). The loading assumption also
fails its own check: the implied $M$ per dose must have log-log slope $1$
against dose, and \NMSlopeExclOne{} of \NMSlopeTested{} intervals exclude it.
An unconstrained per-dose pool beats a proportional one for \NPrefFree{} of
\NDoseFitted{} constructs on both AIC and held-out genes
(\S\ref{sec:convex}). We therefore report no estimate of $\rho$: the
inverted values span \RhoDataDecades{} decades across construct-dose cells,
leaving the $\alpha$-driven uncertainty of Figure~\ref{fig:phase} where it
was.

\section{Limitations and conclusion}

For within-construct fixed-dose rank statistics of fractional occupancy under
the one-effective-$K$ model, reweighting $K$ cannot outperform the $K$ it is
given: Proposition~\ref{prop:rank} says no such statistic detects the coupling
at all. That is the most useful thing the shared pool buys, because it says
which experiments are not worth running (\S\ref{sec:limits}). The archival data buy less: a weak association in the right
direction, no established added value from coupling, and dose fits too
heterogeneous to identify $\rho$. The compression of \S\ref{sec:approx} is
accurate but not faster at the size measured. A decisive experiment needs more
doses, constructs that are not variants of one another, and a better $K$.

The next step is a learned, strictly positive affinity head, a transformer or
graph network over base-pairing and accessibility, emitting $K_j>0$ per pair
and trained through the conservation equation to learn what the ViennaRNA
construction omits. Evaluation should hold out guides and cell lines and
compare the head with and without the layer. It is not evaluated here.

\bibliographystyle{plainnat}
\bibliography{refs}

@article{arvey2010,
  author = {Arvey, A and Larsson, E and Sander, C and Leslie, CS and Marks, DS},
  title = {Target {mRNA} abundance dilutes {microRNA} and {siRNA} activity},
  journal = {Molecular Systems Biology},
  year = {2010},
  volume = {6},
  pages = {363},
  doi = {10.1038/msb.2010.24}
}

@article{athar2019,
  author = {Athar, A and F{\"u}llgrabe, A and George, N and Iqbal, H and Huerta, L and Ali, A and Snow, C and Fonseca, NA and Petryszak, R and Papatheodorou, I and Sarkans, U and Brazma, A},
  title = {{ArrayExpress} update: from bulk to single-cell expression data},
  journal = {Nucleic Acids Research},
  year = {2019},
  volume = {47},
  number = {D1},
  pages = {D711-D715},
  doi = {10.1093/nar/gky964}
}

@article{bai2024,
  author = {Bai, Y and Zhong, H and Wang, T and Lu, ZJ},
  title = {{OligoFormer}: an accurate and robust prediction method for {siRNA} design},
  journal = {Bioinformatics},
  year = {2024},
  volume = {40},
  number = {10},
  pages = {btae577},
  doi = {10.1093/bioinformatics/btae577}
}

@article{barrett2013,
  author = {Barrett, T and Wilhite, SE and Ledoux, P and Evangelista, C and Kim, IF and Tomashevsky, M and Marshall, KA and Phillippy, KH and Sherman, PM and Holko, M and Yefanov, A and Lee, H and Zhang, N and Robertson, CL and Serova, N and Davis, S and Soboleva, A},
  title = {{NCBI} {GEO}: archive for functional genomics data sets---update},
  journal = {Nucleic Acids Research},
  year = {2013},
  volume = {41},
  number = {D1},
  pages = {D991-D995},
  doi = {10.1093/nar/gks1193}
}

@article{bernhart2006,
  author = {Bernhart, SH and Hofacker, IL and Stadler, PF},
  title = {Local {RNA} base pairing probabilities in large sequences},
  journal = {Bioinformatics},
  year = {2006},
  volume = {22},
  number = {5},
  pages = {614-615},
  doi = {10.1093/bioinformatics/btk014}
}

@article{birmingham2006,
  author = {Birmingham, A and Anderson, EM and Reynolds, A and Ilsley-Tyree, D and Leake, D and Fedorov, Y and Baskerville, S and Maksimova, E and Robinson, K and Karpilow, J and Marshall, WS and Khvorova, A},
  title = {3' {UTR} seed matches, but not overall identity, are associated with {RNAi} off-targets},
  journal = {Nature Methods},
  year = {2006},
  volume = {3},
  number = {3},
  pages = {199-204},
  doi = {10.1038/nmeth854}
}

@article{bosson2014,
  author = {Bosson, AD and Zamudio, JR and Sharp, PA},
  title = {Endogenous {miRNA} and target concentrations determine susceptibility to potential {ceRNA} competition},
  journal = {Molecular Cell},
  year = {2014},
  volume = {56},
  number = {3},
  pages = {347-359},
  doi = {10.1016/j.molcel.2014.09.018}
}

@article{burchard2009,
  author = {Burchard, J and Jackson, AL and Malkov, V and Needham, RH and Tan, Y and Bartz, SR and Dai, H and Sachs, AB and Linsley, PS},
  title = {{MicroRNA}-like off-target transcript regulation by {siRNAs} is species specific},
  journal = {RNA},
  year = {2009},
  volume = {15},
  number = {2},
  pages = {308-315},
  doi = {10.1261/rna.1326809}
}

@article{caffrey2011,
  author = {Caffrey, DR and Zhao, J and Song, Z and Schaffer, ME and Haney, SA and Subramanian, RR and Seymour, AB and Hughes, JD},
  title = {{siRNA} off-target effects can be reduced at concentrations that match their individual potency},
  journal = {PLoS ONE},
  year = {2011},
  volume = {6},
  number = {7},
  pages = {e21503},
  doi = {10.1371/journal.pone.0021503}
}

@article{dai2005,
  author = {Dai, M and Wang, P and Boyd, AD and Kostov, G and Athey, B and Jones, EG and Bunney, WE and Myers, RM and Speed, TP and Akil, H and Watson, SJ and Meng, F},
  title = {Evolving gene/transcript definitions significantly alter the interpretation of {GeneChip} data},
  journal = {Nucleic Acids Research},
  year = {2005},
  volume = {33},
  number = {20},
  pages = {e175},
  doi = {10.1093/nar/gni179}
}

@article{denzler2014,
  author = {Denzler, R and Agarwal, V and Stefano, J and Bartel, DP and Stoffel, M},
  title = {Assessing the {ceRNA} hypothesis with quantitative measurements of {miRNA} and target abundance},
  journal = {Molecular Cell},
  year = {2014},
  volume = {54},
  number = {5},
  pages = {766-776},
  doi = {10.1016/j.molcel.2014.03.045}
}

@article{denzler2016,
  author = {Denzler, R and McGeary, SE and Title, AC and Agarwal, V and Bartel, DP and Stoffel, M},
  title = {Impact of {microRNA} levels, target-site complementarity, and cooperativity on competing endogenous {RNA}-regulated gene expression},
  journal = {Molecular Cell},
  year = {2016},
  volume = {64},
  number = {3},
  pages = {565-579},
  doi = {10.1016/j.molcel.2016.09.027}
}

@article{frankish2021,
  author = {Frankish, A and Diekhans, M and Jungreis, I and Lagarde, J and Loveland, JE and Mudge, JM and Sisu, C and Wright, JC and Armstrong, J and Barnes, I and Berry, A and Bignell, A and Boix, C and Carbonell Sala, S and Cunningham, F and Di Domenico, T and Donaldson, S and Fiddes, IT and Garc{\'i}a Gir{\'o}n, C and Gonzalez, JM and Grego, T and Hardy, M and Hourlier, T and Howe, KL and Hunt, T and Izuogu, OG and Johnson, R and Martin, FJ and Mart{\'i}nez, L and Mohanan, S and Muir, P and Navarro, FCP and Parker, A and Pei, B and Pozo, F and Riera, FC and Ruffier, M and Schmitt, BM and Stapleton, E and Suner, MM and Sycheva, I and Uszczynska-Ratajczak, B and Wolf, MY and Xu, J and Yang, YT and Yates, A and Zerbino, D and Zhang, Y and Choudhary, JS and Gerstein, M and Guig{\'o}, R and Hubbard, TJP and Kellis, M and Paten, B and Tress, ML and Flicek, P},
  title = {{GENCODE} 2021},
  journal = {Nucleic Acids Research},
  year = {2021},
  volume = {49},
  number = {D1},
  pages = {D916-D923},
  doi = {10.1093/nar/gkaa1087}
}

@article{garcia2011,
  author = {Garcia, DM and Baek, D and Shin, C and Bell, GW and Grimson, A and Bartel, DP},
  title = {Weak seed-pairing stability and high target-site abundance decrease the proficiency of lsy-6 and other {microRNAs}},
  journal = {Nature Structural \& Molecular Biology},
  year = {2011},
  volume = {18},
  number = {10},
  pages = {1139-1146},
  doi = {10.1038/nsmb.2115}
}

@article{huesken2005,
  author = {Huesken, D and Lange, J and Mickanin, C and Weiler, J and Asselbergs, F and Warner, J and Meloon, B and Engel, S and Rosenberg, A and Cohen, D and Labow, M and Reinhardt, M and Natt, F and Hall, J},
  title = {Design of a genome-wide {siRNA} library using an artificial neural network},
  journal = {Nature Biotechnology},
  year = {2005},
  volume = {23},
  number = {8},
  pages = {995-1001},
  doi = {10.1038/nbt1118}
}

@article{irizarry2003,
  author = {Irizarry, RA and Hobbs, B and Collin, F and Beazer-Barclay, YD and Antonellis, KJ and Scherf, U and Speed, TP},
  title = {Exploration, normalization, and summaries of high density oligonucleotide array probe level data},
  journal = {Biostatistics},
  year = {2003},
  volume = {4},
  number = {2},
  pages = {249-264},
  doi = {10.1093/biostatistics/4.2.249}
}

@article{jackson2003,
  author = {Jackson, AL and Bartz, SR and Schelter, J and Kobayashi, SV and Burchard, J and Mao, M and Li, B and Cavet, G and Linsley, PS},
  title = {Expression profiling reveals off-target gene regulation by {RNAi}},
  journal = {Nature Biotechnology},
  year = {2003},
  volume = {21},
  number = {6},
  pages = {635-637},
  doi = {10.1038/nbt831}
}

@article{jackson2006seed,
  author = {Jackson, AL and Burchard, J and Schelter, J and Chau, BN and Cleary, M and Lim, L and Linsley, PS},
  title = {Widespread {siRNA} ``off-target'' transcript silencing mediated by seed region sequence complementarity},
  journal = {RNA},
  year = {2006},
  volume = {12},
  number = {7},
  pages = {1179-1187},
  doi = {10.1261/rna.25706}
}

@article{janas2012,
  author = {Janas, MM and Wang, B and Harris, AS and Aguiar, M and Shaffer, JM and Subrahmanyam, YV and Behlke, MA and Wucherpfennig, KW and Gygi, SP and Gagnon, E and Novina, CD},
  title = {Alternative {RISC} assembly: binding and repression of {microRNA}-{mRNA} duplexes by human {Ago} proteins},
  journal = {RNA},
  year = {2012},
  volume = {18},
  number = {11},
  pages = {2041-2055},
  doi = {10.1261/rna.035675.112}
}

@article{khan2009,
  author = {Khan, AA and Betel, D and Miller, ML and Sander, C and Leslie, CS and Marks, DS},
  title = {Transfection of small {RNAs} globally perturbs gene regulation by endogenous {microRNAs}},
  journal = {Nature Biotechnology},
  year = {2009},
  volume = {27},
  number = {6},
  pages = {549-555},
  doi = {10.1038/nbt.1543}
}

@article{loinger2012,
  author = {Loinger, A and Shemla, Y and Simon, I and Margalit, H and Biham, O},
  title = {Competition between small {RNAs}: a quantitative view},
  journal = {Biophysical Journal},
  year = {2012},
  volume = {102},
  number = {8},
  pages = {1712-1721},
  doi = {10.1016/j.bpj.2012.01.058}
}

@article{lorenz2011,
  author = {Lorenz, R and Bernhart, SH and H{\"o}ner Zu Siederdissen, C and Tafer, H and Flamm, C and Stadler, PF and Hofacker, IL},
  title = {{ViennaRNA} {Package} 2.0},
  journal = {Algorithms for Molecular Biology},
  year = {2011},
  volume = {6},
  pages = {26},
  doi = {10.1186/1748-7188-6-26}
}

@article{marinov2014,
  author = {Marinov, GK and Williams, BA and McCue, K and Schroth, GP and Gertz, J and Myers, RM and Wold, BJ},
  title = {From single-cell to cell-pool transcriptomes: stochasticity in gene expression and {RNA} splicing},
  journal = {Genome Research},
  year = {2014},
  volume = {24},
  number = {3},
  pages = {496-510},
  doi = {10.1101/gr.161034.113}
}

@article{riba2017,
  author = {Riba, A and Emmenlauer, M and Chen, A and Sigoillot, F and Cong, F and Dehio, C and Jenkins, J and Zavolan, M},
  title = {Explicit modeling of {siRNA}-dependent on- and off-target repression improves the interpretation of screening results},
  journal = {Cell Systems},
  year = {2017},
  volume = {4},
  number = {2},
  pages = {182-193.e4},
  doi = {10.1016/j.cels.2017.01.011}
}

@article{sayers2026,
  author = {Sayers, EW and Bolton, EE and Fine, AM and Kelly, C and Kim, S and Landrum, M and Lathrop, S and Malheiro, A and Murphy, TD and Phan, L and Pujar, S and Trawick, BW and Schneider, VA and Pruitt, KD},
  title = {Database resources of the {National} {Center} for {Biotechnology} {Information} in 2026},
  journal = {Nucleic Acids Research},
  year = {2026},
  volume = {54},
  number = {D1},
  pages = {D20-D27},
  doi = {10.1093/nar/gkaf1060}
}

@article{wang2012,
  author = {Wang, D and Zhang, Z and O'Loughlin, E and Lee, T and Houel, S and O'Carroll, D and Tarakhovsky, A and Ahn, NG and Yi, R},
  title = {Quantitative functions of {Argonaute} proteins in mammalian development},
  journal = {Genes \& Development},
  year = {2012},
  volume = {26},
  number = {7},
  pages = {693-704},
  doi = {10.1101/gad.182758.111}
}

@article{wee2012,
  author = {Wee, LM and Flores-Jasso, CF and Salomon, WE and Zamore, PD},
  title = {{Argonaute} divides its {RNA} guide into domains with distinct functions and {RNA}-binding properties},
  journal = {Cell},
  year = {2012},
  volume = {151},
  number = {5},
  pages = {1055-1067},
  doi = {10.1016/j.cell.2012.10.036}
}

\clearpage
\appendix

\section{Pipeline overview}
\label{sec:workflow}

\begin{figure}[H]
\centering
\includegraphics[width=\textwidth]{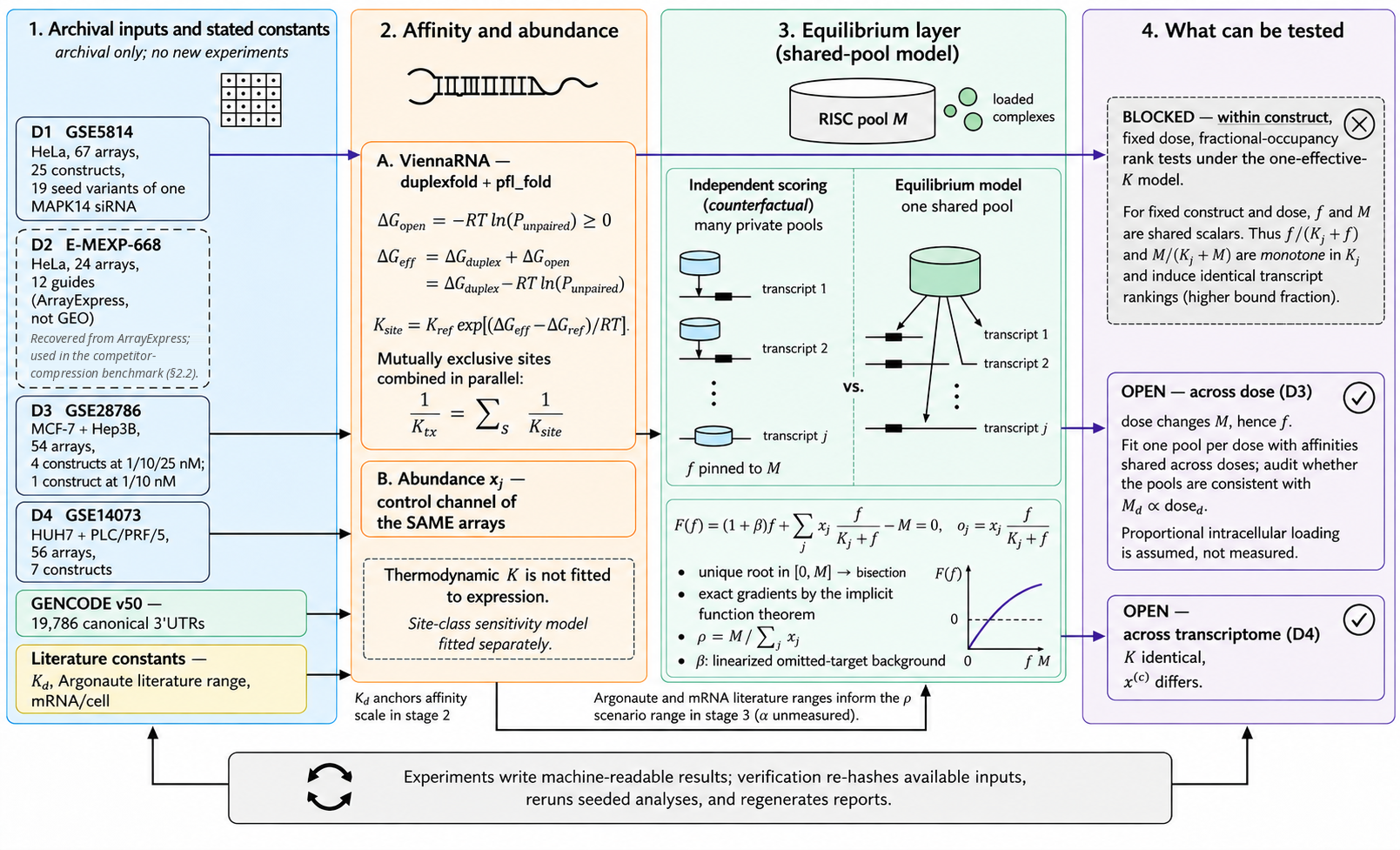}
\caption{The full pipeline, shown at page width because the panel text is
part of the content. Archival expression datasets, canonical 3$'$UTR
sequences, computed thermodynamic affinities, and stated literature ranges
are combined in a differentiable shared-pool equilibrium model. At fixed
construct and dose, equilibrium and independent fractional-occupancy scores
induce identical transcript rankings under the one-effective-$K$ model. Pool
effects instead remain testable across doses or transcriptomic contexts.
Intracellular loading proportional to transfected dose is an explicit
modelling assumption rather than a direct measurement. Boxes name the
archival accessions, the affinity construction of \S\ref{sec:scale}, the
conservation equation \eqref{eq:cons} of the main text, and which
experiments each part licenses.}
\label{fig:workflow}
\end{figure}

Figure~\ref{fig:workflow} is the whole study on one page, in four stages.
\textbf{Stage 1} is what enters: four archival off-target microarray
accessions, GENCODE canonical 3$'$UTRs, and two literature constants: the
Argonaute copy number, carried as a range, and the mRNA count per cell, a
single value inside the range its source reports (\S\ref{sec:limits}). No new measurement is made anywhere in this work.
\textbf{Stage 2} builds the two per-transcript quantities the layer needs.
Abundance $x_j$ comes from the control measurement of the \emph{same}
experiment that supplies the measured response, so abundance and response are never taken from
different experiments. Affinity $K_j$ comes from a ViennaRNA construction,
duplex energy plus the cost of opening the site, anchored on a measured
seed-match $K_d$ (\S\ref{sec:scale}); it is never fitted to expression, which
is what keeps the association tests in \S\ref{sec:assoc} honest.
\textbf{Stage 3} is the conservation equation \eqref{eq:cons} itself, beside
the independent counterfactual it is compared against, with the properties
that make it usable as a layer: a unique root, bisection, and exact implicit
gradients. \textbf{Stage 4} records what this construction can and cannot
test. The within-construct fixed-dose comparison is blocked by
Proposition~\ref{prop:rank}, not by a limitation of the data; the dose and
cross-transcriptome comparisons remain open, which is why they are the two
experiments attempted.

\section{Code and data availability}

The equilibrium layer, the acquisition scripts, the per-experiment result
files and the figure code will be released publicly once review is complete.
Every number in this manuscript, including every number in a caption, is
generated from a machine-readable result file rather than transcribed; each
generated macro carries the expression and the result file it was read from,
so the trace from a printed number back to the script that produced it is
mechanical. Every retrieved byte carries a provenance record with its URL,
checksum and retrieval status, failed retrievals included. The released
materials also contain a held-out test on external families of a site-class
correction to $K$, which reduced performance and is not reported here. All datasets used
are archival and already public under the accessions named in
\S\ref{sec:data}.

\begin{figure}[h]
\centering
\includegraphics[width=\textwidth]{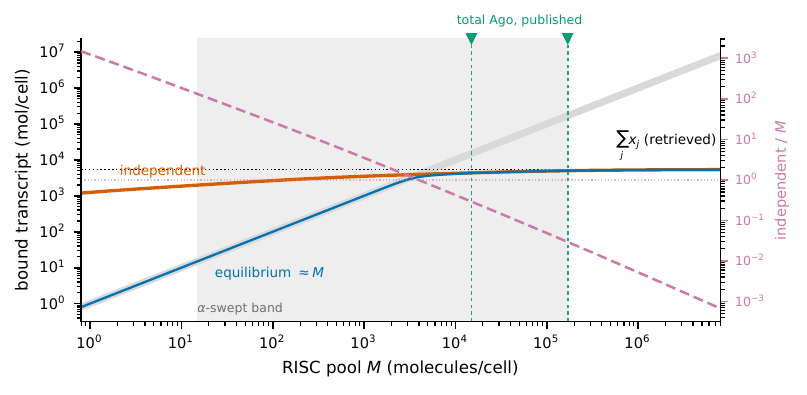}
\caption{Model calculation on measured inputs (abundances from the GSE5814
mock channel, structure from GENCODE v\GencodeRelease{} 3'UTRs; MAPK14-193
parent). Total bound transcript against the loaded pool, with the
over-allocation ratio on the right axis: the independent counterfactual
exceeds the budget by up to \BudgetWorst-fold more complexes, but only at
pools far below one complex per cell, while equilibrium is constrained to the
budget by construction. Green triangles are published Argonaute copy numbers,
total Ago1--4 in the HeLa case; the grey band is the swept loaded fraction. Redistribution and the
high-resource limit are shown in \S\ref{sec:proofs}.}
\label{fig:mech}
\end{figure}

\begin{figure}[h]
\centering
\includegraphics[width=0.58\textwidth]{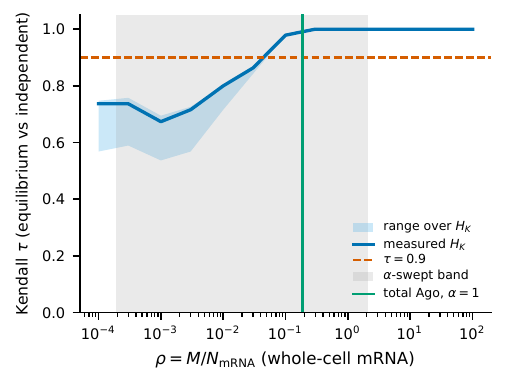}
\caption{Model calculation on measured GSE5814 HeLa abundances. Kendall
$\tau$ between the two scorings' candidate rankings against $\rho$, with the
band covering the swept $H_K$. The $\tau=0.9$ contour lies \emph{inside} the
\RhoDecades-decade scenario range, so the calculation cannot say which side a
real cell falls on: the identifiability problem \S\ref{sec:dose} does not
close.}
\label{fig:phase}
\end{figure}

\section{Proofs}
\label{sec:proofs}

\begin{figure}[h]
\centering
\includegraphics[width=\textwidth]{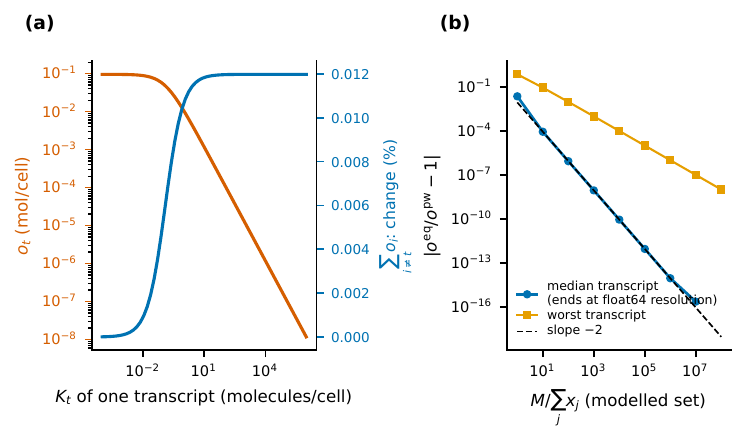}
\caption{Model calculation on measured GSE5814 abundances and GENCODE
v\GencodeRelease{} structure. (a) Redistribution: raising one
transcript's $K_t$ lowers its own occupancy and raises every other's,
monotonically. (b) The relative gap between the two scorings decays with
exponent $\PwSlope \pm \PwSlopeSE$ against a predicted $\PwPredicted$, over
\PwDecades{} decades.}
\label{fig:theory}
\end{figure}

\paragraph{Gradient accuracy.} Over \NGradCoord{} coordinates the median
relative disagreement with a central difference is \GradErr{}. The worst is
\GradWorst{}, and it is not informative: it falls on a coordinate whose own
gradient is near zero, where a relative measure divides by nothing. Measured
against the gradient norm, which is the quantity that says whether the
implicit gradients are right, the worst error is \GradWorstNorm{}. A wrong
implicit gradient would disagree by order one, not by parts in $10^7$.

\paragraph{Redistribution.} Assume $M>0$, $x_j\ge 0$, $K_j>0$ and
$\beta\ge 0$, which are the conditions under which the root of
\eqref{eq:cons} is unique. Write
\[
D \;=\; \frac{\partial F}{\partial f}
\;=\; (1+\beta) + \sum_j \frac{x_j K_j}{(K_j+f)^2} \;>\; 0 .
\]
The implicit function theorem gives
$\partial f/\partial K_t = x_t f / [(K_t+f)^2 D] > 0$: raising one
transcript's $K_t$ raises the free pool. For any other transcript $i\neq t$,
\[
\frac{\partial o_i}{\partial K_t}
= \frac{x_i K_i}{(K_i+f)^2}\,\frac{\partial f}{\partial K_t} \;>\; 0 ,
\]
so every other transcript gains. For the swept transcript itself,
$\partial o_t/\partial K_t = x_t(f'K_t-f)/(K_t+f)^2$ where
$f' = \partial f/\partial K_t$; putting $u = x_t K_t/(K_t+f)^2$ gives
$f'K_t - f = f(u/D - 1)$, which is strictly negative because
$D \ge 1+u > u$. So $o_t$ falls while every $o_i$ rises.

\section{Binned background at shallow retrieval}
\label{sec:binned}

\begin{table}[h]
\caption{Maximum relative error in on-target occupancy against the full
\EFourN-transcript reference, in per cent, over $\rho\in\{0.01,0.5,10\}$,
quantile bins. The first column is the linear affinity-weighted rule, which
uses no bins at all. The columns to its right are $B$ \emph{saturable} bins,
so the $B{=}1$ column is one saturable bin: a different object, which errs
differently. Forcing a saturable bin to its linear limit does reproduce the
linear column exactly, to a maximum relative difference of \RegDiff{}, but
that is a regression check on the implementation and not a row of this
table.}
\label{tab:e4b}
\centering
\small
\begin{tabular}{rrrrrrrr}
\toprule
& linear & \multicolumn{6}{c}{saturable bins, $B=$} \\
\cmidrule(lr){3-8}
$R$ & (no bins) & 1 & 2 & 5 & 10 & 50 & 200 \\
\midrule
25 & 91.1 & 74.6 & 72.5 & 64.6 & 27.1 & 1.13 & 0.0548 \\
100 & 26.7 & 8.21 & 7.88 & 6.13 & 3.06 & 0.153 & 0.0137 \\
500 & 0.0233 & 0.00155 & 0.000502 & 9.22e-05 & 2.25e-05 & 2.24e-06 & 1.32e-06 \\
\bottomrule
\end{tabular}
\end{table}

\paragraph{How much does binning actually save?} Restoring $1\%$ accuracy at
$R=25$ takes $25+\BSmallestOne$ individually scored terms, and the two binning
rules agree exactly on that count, against the $501$ an untruncated $R=500$
requires. That is not a saving in time: \S\ref{sec:approx} measures the
compressed solve as slower than the full one at this size, so a smaller term
count must not be read as a speed-up. The table reports quantile bins,
which are \QuantileCheaperClause{} across the \NBinSettings{} depth and
threshold settings swept, so nothing here is flattered by the choice.

\section{Compressed solve: selection, gradients and timing}
\label{sec:apxdetail}

\begin{figure}[h]
\centering
\includegraphics[width=\textwidth]{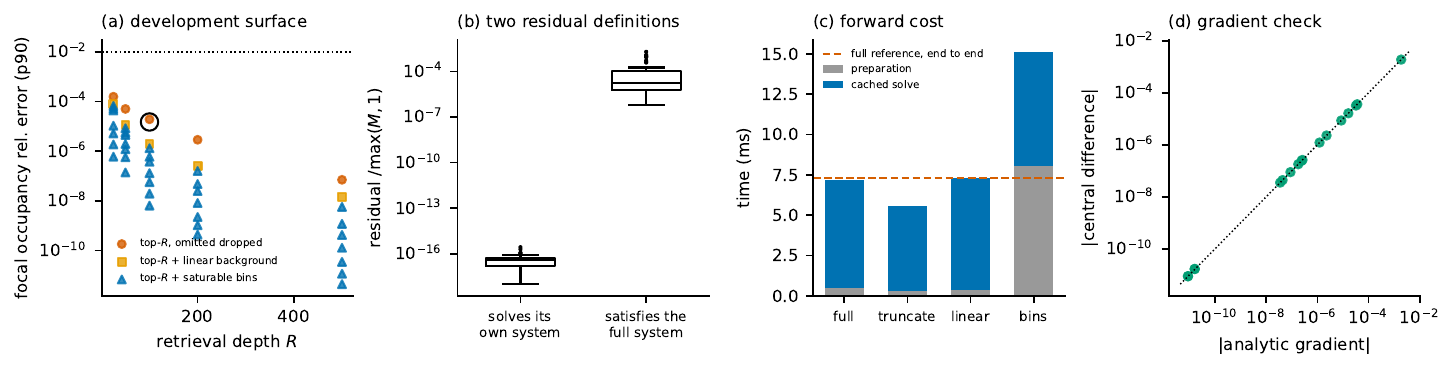}
\caption{Detail behind Figure~\ref{fig:approx}. (a) Development selection
surface: 90th-percentile focal occupancy error over the \ApxDevFam{}
development families for every configuration on the frozen grid; the circled
point is the selected one, fixed before any held-out family was scored.
(b) The two residual definitions on the \ApxCases{} held-out cases, each
divided by $\max(M,1)$. (c) Median forward cost at matched outputs,
preparation and cached solve stacked, against the full reference end to end.
(d) Analytic gradients against central differences over \ApxFDCoord{}
coordinates including omitted competitors; the dotted line is equality.}
\label{fig:apxdetail}
\end{figure}

\paragraph{Selection.} The grid over method, retrieval depth $R$ and bin count
$B$ was scored on \ApxDevFam{} guide families, three of them D1 target genes
and the rest D2 seed families. The configuration
(\ApxMethod{}, $R=\ApxR{}$, $B=\ApxB{}$) was frozen there and applied
unchanged to \ApxTestFam{} disjoint families, \ApxCases{} cases, which are
\ApxTestCon{} constructs times \ApxRhos{} values of $\rho$; the families are
the independent units, not the cases. No configuration was reselected after
the held-out families were scored.

\paragraph{Gradients.} Central differences were taken on \ApxFDCoord{}
coordinates, including coordinates of omitted competitors, with the step
escalated until the difference cleared the float64 noise floor. The worst
relative error is \ApxFDWorst{}; restricted to the \ApxFDAbove{} coordinates
whose analytic gradient exceeds $10^{-9}$, below which a relative measure
divides by the noise floor, it is \ApxFDStrat{}. Both numbers are reported
because neither replaces the other.

\paragraph{Timing.} Preparation, bin construction, cached solve and end-to-end
cost are measured separately on one construct with \ApxNRef{} competitors,
CPU, float64, single-threaded, \ApxReps{} timed repetitions after warm-up,
with every method returning the same outputs. Bin construction is a component
of preparation, not an addition to it. The solver runs a fixed \ApxIters{}
bisection iterations for every method, so iteration count cannot separate
them; what does is the number of reduction passes per iteration. Adding
\ApxElems{} elements to an existing reduction costs \ApxCostElems\,ms, while
adding a second reduction holding a single element costs
\ApxCostSecond\,ms, so per-pass fixed cost dominates the $O(n)$ work at this
size and the linear and binned backgrounds, which each add a pass, cannot
beat the full solve. Charging preparation to both sides, truncation is
cheaper from the first solve and the binned configuration never breaks even.
These are measurements of this workload on this machine, not scaling claims.

\section{The site-class hierarchy is not the canonical one}
\label{sec:hier}

\begin{figure}[h]
\centering
\includegraphics[width=0.56\textwidth]{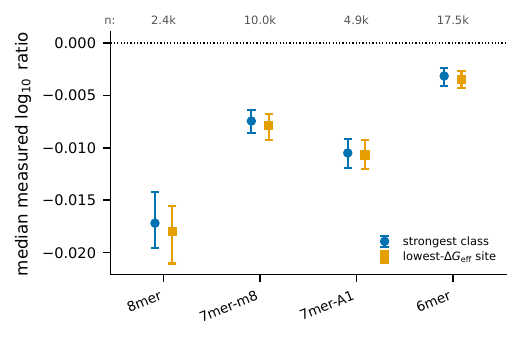}
\caption{Measured repression by seed site class over \NullConstructs{}
GSE5814 constructs under both site-assignment rules, bootstrap 95\% intervals
on the median, pair counts above each pair. The two 7mer classes are
exchanged relative to the canonical order under both rules, which is the
observation this section is about.}
\label{fig:hier}
\end{figure}

Seed-matched transcripts are repressed against transcripts with no site, but
the four classes do not order as the canonical hierarchy expects. Pooling
\NullConstructs{} constructs and assigning each transcript the strongest class
present on it, the medians are \CanEightMerMed{} for 8mer (n = \CanEightMerN,
95\% CI \CanEightMerCI), \CanSevenMEightMed{} for 7mer-m8 (n = \CanSevenMEightN,
\CanSevenMEightCI), \CanSevenAOneMed{} for 7mer-A1 (n = \CanSevenAOneN,
\CanSevenAOneCI) and \CanSixMerMed{} for 6mer (n = \CanSixMerN, \CanSixMerCI).
That sequence is \CanMonotone{} in the canonical order 8mer, 7mer-m8,
7mer-A1, 6mer: the two 7mer classes are exchanged, with 7mer-A1 the more
repressed of the two by \CanInv{} (95\% CI \CanInvCI), and the exchange
appears in \CanNInv{} of \CanNTested{} constructs individually.

It is not an artefact of how a transcript is assigned a class. Under the
alternative rule, labelling each transcript by the class of its lowest
$\Delta G_{\mathrm{eff}}$ site, the ordering is \BestMonotone{} in the same way and the difference is
\BestInv{} (95\% CI \BestInvCI) in \BestNInv{} of \BestNTested{} constructs.
Both intervals exclude zero. This is not an anomaly to be explained away. A
7mer-A1 site pairs guide positions 2--7 and adds the $t1$ adenosine, which
Argonaute reads through a dedicated pocket rather than by base pairing; a
7mer-m8 site pairs 2--8 and has no $t1$. An ordering in which 7mer-A1 wins is
what a strong $t1$ contribution looks like. It is consistent with a
contribution from the Argonaute $t1$-adenosine pocket, the same mechanism
invoked in \S\ref{sec:assoc} for why a duplex-energy model ranks these
transcripts poorly: the term that separates the two classes is not a base
pair, and the present duplex-pairing-only energy construction does not
represent this non-base-pair contribution.

\section{How the association is tested}
\label{sec:paired}

\paragraph{The permutation null is not centred on zero.} Permuting measured
values within construct preserves each construct's response distribution, so
any between-construct structure survives the shuffle: pooling constructs whose
predicted occupancies and whose measured responses both differ systematically
leaves a correlation the shuffle cannot remove. The null is therefore reported
rather than assumed, and the test is two-sided about the null's own mean
rather than about zero. The two scorings do not share that mean, which is why
a raw difference of pooled correlations is not a test: at $\rho=\GapRho{}$
the equilibrium null sits at one place and the independent null at another,
and the difference between the nulls, \GapNullMean{}, exceeds the raw gap
\SpearGap{} in magnitude, which is why the net changes sign rather than
merely shrinking. Net of their own nulls the two associations are \NullCentredEq{}
and \NullCentredInd{}.

\paragraph{The paired test, and the criterion it feeds.} One within-construct
permutation is drawn and applied to \emph{both} models, and the difference of
the two correlations is recorded under that same shuffle, so the two models
see identical permuted data and the difference is not inflated by their
seeing different ones. It does \emph{not} follow that the confound cancels:
the paired null mean is \GapNullMean{}, not zero, which is why the test is
two-sided about that mean rather than about zero. Over \NullPerm{} draws this
gives the paired null quoted in \S\ref{sec:assoc}.

The screen is prespecified and has three parts, all of which must hold at the
same $\rho$: the paired permutation $p$ is below $0.05$, the
construct-cluster interval on the gap excludes zero, and the gap net of the
paired null is in equilibrium's favour. That is what the count in
\S\ref{sec:assoc} reports. A fourth requirement separates a difference that is
detectable from one worth having: the net gap must be at least one
construct-cluster standard error of the equilibrium association itself. The
scale is taken from the data rather than chosen, and the fourth requirement is
met at no $\rho$, which is why no added predictive value is claimed.

\paragraph{Two bootstraps, and what each licenses.} The pair-level bootstrap
resamples transcript-construct pairs. It is \emph{conditional on the observed
constructs}: it describes sampling noise inside this panel of
\NullConstructs{} constructs and cannot support a statement about a new guide.
The construct-cluster bootstrap resamples whole constructs with replacement,
keeping every pair belonging to a sampled construct, over \ClusterBoot{}
draws; that is the interval a generalising claim needs, and it is
correspondingly wider. At $\rho=\GapRho{}$ the pair-level interval on the
equilibrium correlation is \SpearBestPairCI{} and the cluster interval on the
raw gap is \GapClusterCI{}. The latter excludes zero, but it is an interval on
the \emph{raw} gap, which carries the null baseline with it, so it is not the
quantity the claim rests on; the null-centred difference is what is, and that
is \GapNet{}. Neither removes the dependence created by one gene
appearing under many constructs: the \NullNPairs{} pooled pairs cover
\NGeneRepeat{} distinct genes, the most frequent appearing under
\MaxConstructsPerGene{} of them. The cluster bootstrap carries a gene's
repeated appearances together with the construct that produced them, but
residual gene-level dependence across distinct constructs would need a
crossed random-effects model that \NullConstructs{} constructs cannot
support. We therefore treat \S\ref{sec:assoc} as an upper bound on what can
be claimed, not a lower one.

\section{Estimator validation}
\label{sec:sim}

The site-class estimator pins the 7mer-m8 class to the published
dissociation constant \citep{wee2012} to fix the common scale, and its
recovered class ordering is a built-in check: it comes out correct (8mer
strongest) for \ClassSane{} of \NDoseFitted{} constructs.

SIMULATION. Ground-truth pools and affinities do not exist in nature, so this
validates the estimators of \S\ref{sec:dose} and claims nothing about any
cell. Data
were generated from each model in turn at noise levels \SimNoise{} and refitted
blind. On data generated by independent scoring, where the true exponent is
exactly 1, the fitted exponent is correct to within \SimSlopeErrInd. On data
generated by the equilibrium model, whose own exponent is \SimTrueEqSlope{}
and therefore above 1 as the convexity argument requires, the fitted exponent
is correct to within \SimSlopeErrEq. The site-class estimator recovers its
true exponent to within \SimSlopeErrClass, the true pools to within
\SimPoolErr\% and the true per-class affinities to within \SimKErr\%, with the
class ordering recovered correctly at every noise level. The simulation
verifies implementation-level recovery under the assumed generative model; it
does not rule out model misspecification, confounding, or weak
identifiability in the archival data, and the audit in \S\ref{sec:dose}
finds all three to be live.

\begin{table}[h]
\caption{Dose series, per construct. Exponent is
$\mathrm{d}\log(\text{pool})/\mathrm{d}\log(\text{dose})$ under the site-class
estimator with a bootstrap interval over transcripts; $1$ is the
independent-scoring prediction and below $1$ is incompatible with the combined
model under proportional loading. The
sample sizes belong to different estimators: $n$ thermo.\ is the retrieved
subset, $n$ class every measured gene. $M$ slope must be $1$ if proportional
loading holds. $\dagger$: two dose points, zero residual df.}
\label{tab:e9}
\centering
\small
\begin{tabular}{llrrrrrr}
\toprule
construct & cells & doses & $n$ thermo. & $n$ class & exponent & 95\% CI & $M$ slope \\
\midrule
HK2-3581 & Hep3B & 3 & 4,633 & 17,317 & 0.21 & [0.13, 0.27] & 0.10 \\
HK2-3581M & Hep3B & 3 & 4,633 & 17,317 & 1.09 & [0.87, 1.41] & 0.80 \\
HK2-4031 & Hep3B & 2$^{\dagger}$ & 5,625 & 17,317 & 7.99 & [3.70, 18.00] & 4.79 \\
STAT3-1676 & MCF7 & 3 & 5,091 & 17,317 & 0.63 & [0.47, 0.86] & 0.37 \\
STAT3-1676M & MCF7 & 3 & 5,091 & 17,317 & 0.60 & [0.37, 0.86] & 0.32 \\
\bottomrule
\end{tabular}
\end{table}

\section{The dose likelihood as implemented}
\label{sec:like}

Transcribed from the implementation rather than restated. For construct $s$,
dose $d$ and transcript $j$,
\[
y_{d,j} \;=\; a_d \;-\; c_s\, b_j(p_d) \;+\; \varepsilon,
\qquad b_j(p) \;=\; \frac{p}{K_j + p},
\]
with $y$ the measured log$_2$ fold change against the zero-dose arrays. The
per-dose intercepts $a_d$ and the amplitude $c_s$ are profiled out in closed
form: centring $y$ and $b$ within each dose removes $a_d$ exactly, leaving a
single-coefficient regression through the origin pooled over doses,
$c_s = -\sum_d \langle \tilde y_d, \tilde b_d\rangle / \sum_d \langle \tilde
b_d, \tilde b_d\rangle$, the sign because repression is negative. Errors are
homoscedastic within a construct and $\sigma$ is concentrated out, so the
objective is the Gaussian profile negative log-likelihood
$\tfrac{1}{2} n \log(\mathrm{RSS}/n)$; the per-dose residual spread is
reported as a diagnostic rather than modelled, because three doses barely
estimate a per-dose variance. Only $\log_{10} p_d$ is optimised numerically,
by Nelder--Mead from several starts.

$c_s$ is shared across doses within a construct, and that sharing is the whole
identifying assumption: it is the efficacy of a bound complex, a property of
the guide and the silencing machinery, so it cannot depend on how much siRNA
was pipetted on. In the weak-binding limit $K_j \gg p_d$ only the product
$c_s p_d$ is identified, so the absolute pool, and hence $\rho$, rests on
curvature, while the \emph{slope} survives the degeneracy.

In the site-class variant the free parameters are $\log_{10} K_c$ for the
three classes other than the anchor, plus one $\log_{10} p_d$ per dose; the
7mer-m8 class is pinned to the published seed-match $K_d$ to fix the common
scale. Sites combine as $1/K_j = \sum_c n_{jc}/K_c$ over the class counts, and
a transcript with no site has $K_j = \infty$ and hence $b_j = 0$ at any pool,
which is what pins the per-dose intercept. The constrained fits replace the
free pool vector by $M_d = \kappa\,d$ with $\kappa = 10^{u} > 0$, taking
$p_d = M_d$ under independent scoring and $p_d = f(M_d)$ under equilibrium,
$f$ solving \eqref{eq:cons}; all three are compared by
$\mathrm{AIC} = n\log(\mathrm{RSS}/n) + 2k$ and by held-out-gene prediction.

\section{Why the pool exponent in \texorpdfstring{$M$}{M} cannot fall below one}
\label{sec:convex}

From \eqref{eq:cons},
$\mathrm{d}^2M/\mathrm{d}f^2 = -\sum_j 2x_jK_j/(K_j+f)^3 < 0$, so $M$ is
strictly concave in $f$ and $f$ is strictly convex in $M$; with $f(0)=0$ the
ratio $f/M$ increases, which gives
$\mathrm{d}\log f/\mathrm{d}\log M = (\mathrm{d}f/\mathrm{d}M)(M/f) \ge 1$.
The competitor set's buffering capacity saturates, so each added unit of
complex is absorbed less than the last and the free pool accelerates.

The bound is on the exponent in $M$, not in dose. Under the
proportional-loading assumption $M \propto$ dose it carries over, giving
$\mathrm{d}\log f/\mathrm{d}\log(\text{dose}) \ge 1$ for equilibrium against
exactly $1$ for independent scoring. Without it,
\[
\frac{\mathrm{d}\log f}{\mathrm{d}\log(\text{dose})}
= \frac{\mathrm{d}\log f}{\mathrm{d}\log M}\,
  \frac{\mathrm{d}\log M}{\mathrm{d}\log(\text{dose})},
\]
so a measured exponent below $1$ does not by itself falsify equilibrium. It
diagnoses failure of at least one element of the combined specification: the
proportional loading, the affinity or observation model, or the static
equilibrium approximation. \S\ref{sec:dose} finds the first of these failing
its own check, which is why the exponent is reported as an audit rather than
as a measurement of competition.

What is true in the sublinear intuition is a statement about the
\emph{level} of the pool, and it bounds $f$ on both sides:
\[
\frac{M}{\,1+\beta+\sum_j x_j/K_j\,} \;\le\; f(M) \;\le\; \frac{M}{1+\beta}.
\]
The lower bound follows from $f/(K_j+f)\le f/K_j$, which makes the left-hand
side of \eqref{eq:cons} at most $(1+\beta+\sum_j x_j/K_j)f$. The upper bound
follows because every occupancy term is nonnegative, so the left-hand side is
at least $(1+\beta)f$. Since $f$ is convex with $f(0)=0$ it lies above its
tangent at the origin, which is the lower bound. Superlinear elasticity does
not put $f$ above $M/(1+\beta)$: a level bounded by lines through the origin
and an exponent below $1$ are different statements, and only the first
holds.

\section{Approach to the independent limit}
\label{sec:limit}

The result is conditional and the conditions are load-bearing. Assume
$\beta=0$, $K_j$ fixed, $x_j$ fixed, and let $M\to\infty$. Then
$f-(M-\sum_j x_j)\to 0$, so $f\to M-S_{\mathcal{R}}$ and not $M$, and the
difference in \emph{fractional} occupancy between the two scorings is
$O(M^{-2})$ \emph{pointwise for each fixed $K_j$}. Three qualifications
matter. The error over the worst transcript, and separately the error in the
aggregate load, are two different quantities and each decays only as
$O(M^{-1})$ over practical ranges, because the tail of transcripts with
$K_j\gg M$ is not yet in the asymptotic regime at any finite $M$. For fixed $\beta>0$ the
difference from the independent counterfactual is $O(M^{-1})$ in general,
since the background absorbs a constant fraction of the pool. And nothing
here is an unconditional $O(\rho^{-2})$ statement: $\rho$ and $M$ differ by
the fixed abundance total, so the exponent transfers only under the
assumptions above.

Over \PwDecades{} decades on the real transcriptome the median transcript
gives a log-log slope $\PwSlope \pm \PwSlopeSE$ against the predicted
$\PwPredicted$. Both slower quantities land near $-1$: the worst transcript,
which is what Figure~\ref{fig:theory}b plots beside the median, gives
$\PwSlopeMax \pm \PwSlopeMaxSE$, and the aggregate load gives
$\PwSlopeAll \pm \PwSlopeAllSE$. The original derivation stated neither.

\begin{figure}[h]
\centering
\includegraphics[width=\textwidth]{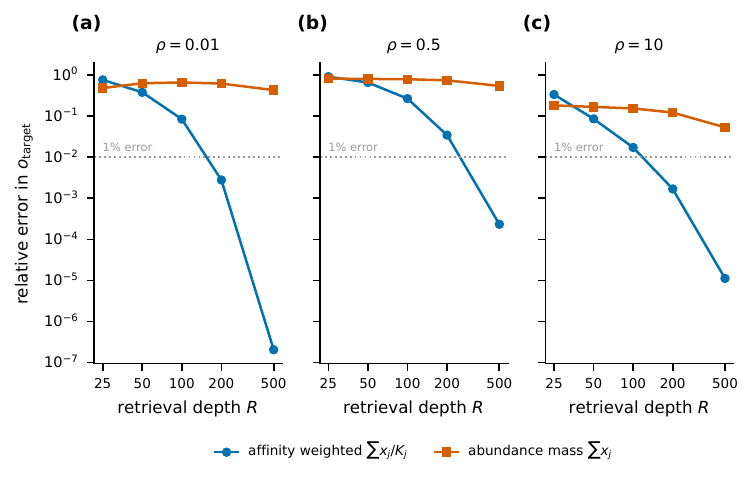}
\caption{Approximation error on data-derived inputs. Relative error in
on-target
occupancy against the full \EFourN-transcript reference when the competitor
set is truncated to the top $R$, under the affinity-weighted calibration rule
and the abundance-mass control, at three values of the competition parameter.
The dotted line is 1\% error.}
\label{fig:retrieval}
\end{figure}

\begin{figure}[h]
\centering
\includegraphics[width=0.49\textwidth]{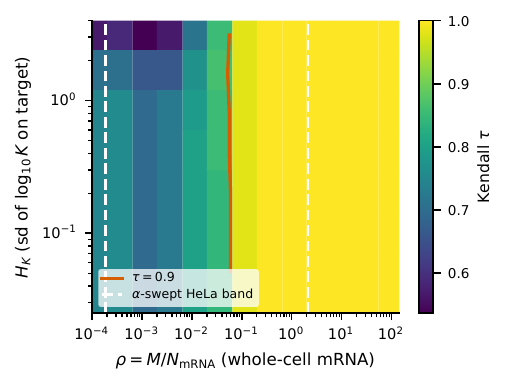}
\hfill
\includegraphics[width=0.49\textwidth]{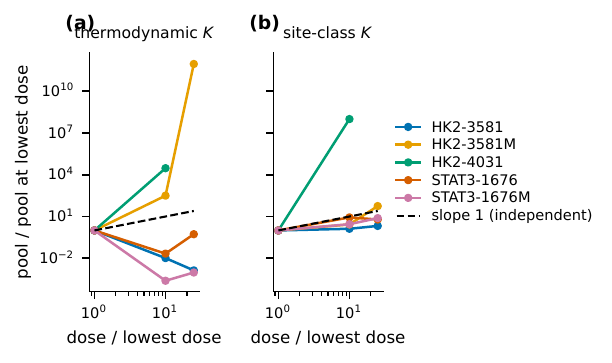}
\caption{Left: the two-dimensional form of Figure~\ref{fig:phase},
kept for completeness; the $\tau = 0.9$ contour runs near-vertical, which is
why the main text shows a one-dimensional curve with an envelope. Right: the
dose fit with each construct's pool normalised to its own lowest dose, under
both affinity models.}
\label{fig:appfigs}
\end{figure}

\section{The cross-context test (exploratory)}
\label{sec:ctx}

Proposition~\ref{prop:rank} permits a second escape: fix the guide and change
the cell, so $K_j$ is identical and only $x^{(c)}$ differs. On
\citet{burchard2009} the pooled ratio of fitted pools between HUH7 and
PLC/PRF/5 is \CtxRatio{} (95\% CI \CtxRatioCI), with \CtxExcl{} of
\CtxAgree{} intervals excluding $1$. We report this as exploratory and draw no
conclusion from it, for three reasons. The seeds are recovered from the same
expression responses used to evaluate the fit, so the analysis is circular;
only constructs whose recovered seeds agree across both lines are retained,
which selects on the outcome; and a ratio of $1$ is predicted only when the
loaded complex is equal in both cell lines, which is unverified: the
on-target readouts differ by \CtxDeliveryMin{}--\CtxDeliveryMax{} log$_2$
units, but that difference is itself a joint consequence of loading,
transcriptome competition and the observation model, so it cannot be read as
evidence of unequal loading. The experiment cannot separate the three.

\citet{burchard2009} profile six APOB siRNAs and a RAD18 control in HUH7 and
PLC/PRF/5. The guide sequences were not included in the GEO deposit, and we did not
locate them in the article or its accessible supplementary material, so the
7-mer site is recovered from the measured response
separately in each cell line by the enrichment scan used for D1, and a
construct enters only when the two recoveries agree exactly; \CtxAgree{} of
\CtxAttempted{} survive. Recovery identifies guide positions 2--8 and nothing
else, so $K$ here is built from the seed duplex rather than the full guide,
which costs resolution within the retrieved set but does not threaten a
comparison that needs only $K$ to be identical between contexts. One effective
pool is fitted per context with a shared amplitude, exactly as for dose.

\section{The affinity scale}
\label{sec:scale}

Per-site free energies are converted to molecules per cell against a
published reference,
$K_{\mathrm{site}} = K_{\mathrm{ref}}\exp[(\Delta G_{\mathrm{eff}} -
\Delta G_{\mathrm{ref}})/RT]$, with $\Delta G_{\mathrm{ref}}$ the median
7mer-m8 site and $K_{\mathrm{ref}}$ the measured seed-match $K_d$
\citep{wee2012} at an assumed cell volume. Sites on one transcript are
parallel binding opportunities, $1/K_j = \sum_s 1/K_{js}$. The competition
parameter uses the whole-cell mRNA total in its denominator, not the retrieved
subset: every experiment sets $M = \rho \cdot(\text{total mRNA})$, so a
$\rho$ quoted here is roughly an order of magnitude smaller than
$M$ divided by the retrieved abundance alone would be. The biochemical
$K_d$ was measured under conditions unlike these cells, so this fixes a
common scale rather than an absolute one. Rank statistics computed
\emph{within} a construct are invariant to it, because rescaling every $K_j$
leaves $q_j$ a monotone function of $K_j$; the pooled statistic of
\S\ref{sec:assoc} mixes constructs whose free pools respond differently to
the rescale and is therefore not exactly invariant, and $\rho$ is not
invariant either.

\section{Standing assumptions}
\label{sec:assumptions}

A transcript is represented by one effective binding state, so its sites are
treated as mutually exclusive and it carries at most one complex. Binding is a
static equilibrium snapshot: recycling and catalytic cleavage are not
modelled. Abundance $x_j$ is taken from the untreated, mock or non-targeting
control measurement of the same experiment, the reference channel for the
two-colour arrays and the control-condition arrays for the single-channel
ones, and is not updated as repression proceeds. Structure comes from canonical GENCODE 3$'$UTRs rather
than cell-specific expressed isoforms, so alternative polyadenylation may
change which sites are actually present. Each of these is a live alternative
explanation for weak affinity performance.

\section{Further limitations}
\label{sec:limits}

The layer models competition among complexes already loaded with one guide, so
loading competition and endogenous microRNA occupancy enter only through $M$.
The molecules-per-cell scale rests on an mRNA count measured in neither
dose-series cell line, and the value used is a point inside the range that
measurement reports rather than a figure the source states, so the exponent is
invariant to it but $\rho$ scales inversely with it.
The site-class affinities are fitted on the same data whose dose dependence is
tested, though shared across doses, which is what keeps the exponent a shape
statement rather than a per-dose fit. The two dose-series cell lines differ
between the STAT3 and HK2 arms, so a construct effect and a cell-line effect
cannot be separated. Three dose points per construct, two for one of them,
leave the per-construct exponents wide, and the bootstrap resamples
transcripts rather than doses, so it cannot represent uncertainty about the
dose grid. The cross-context comparison assumes equal total loaded complex in
both lines, which is unverified; the differing on-target readouts do not
settle it, because loading, transcriptome competition and observation
effects are confounded in them.

\section{Datasets and what had to be recovered}
\label{sec:recover}

\paragraph{Numbering.} The main text numbers the four off-target datasets
D1--D4 in order of appearance. The repository numbers every acquisition,
experiments and references alike: D1 GSE5814, D2 E-MEXP-668, D3 GENCODE,
D4 HeLa abundance, D5 the Huesken efficacy set, D6 literature constants,
D7 GSE28786, D8 GSE14073, D9 a full-text retrieval. So main-text D3 and D4
are repository D7 and D8. Result files use the repository numbering
throughout.

\paragraph{Processing and archives.} Expression values are the depositors'
own summaries, which differ by platform: GSE5814 and E-MEXP-668 are two-colour
arrays deposited as within-array log ratios of the siRNA channel to its
control channel, and the single-channel Affymetrix deposits are RMA summaries
\citep{irizarry2003}. GSE28786 is quantified on a custom CDF
whose probeset identifiers are Entrez gene ids \citep{dai2005}, resolved to
symbols through NCBI Gene \citep{sayers2026}. Accessions are retrieved from
GEO \citep{barrett2013}, except E-MEXP-668, which is in ArrayExpress
\citep{athar2019}.

Guide sequences were absent from two of the four deposits. For GSE5814 they
were taken from \citet{garcia2011} joined on GSM accession and corroborated by
data-driven seed recovery. For GSE14073 the guide sequences were not
included in the deposit and we did not locate them in the article or its
accessible supplementary material, so seeds were recovered from the response
alone. For GSE28786 the sequences are in the
article's own table and were parsed from the full-text XML rather than
transcribed. For E-MEXP-668 they are in the deposit's sample file. In GSE28786
the dose recorded in the sample title disagrees with the dose in the raw file
name for exactly one construct; we resolved it from the measurement, by asking
under which labelling the on-target actually falls.

\end{document}